\pdfoutput=1
\documentclass{IEEEtran}

\usepackage{graphicx}  
\usepackage{amsmath,amssymb}
\usepackage{subcaption}
\usepackage{algorithm}
\usepackage{algpseudocode}

\usepackage[colorlinks]{hyperref}

\begin{document}
	
\title{DarwinML: A Graph-based Evolutionary Algorithm\\
  for Automated Machine Learning}

\author{Fei Qi, Zhaohui Xia, Gaoyang Tang, Hang Yang, Yu Song, Guangrui Qian,\\
  Xiong An, Chunhuan Lin, Guangming Shi\\
  \thanks{This work was supported in part by the National Natural Science
    Foundation of China under Grant 61572387.}
  \thanks{F. Qi, Z. Xia, X. An, C. Lin and G. Shi are with the School of
    Artificial Intelligence, Xidian University, Xi'an 710071, China (e-mail:
    fred.qi@ieee.org; considerxzh@qq.com; anxiong1994@163.com; lch\_xdu@163.com;
    gmshi@xidian.edu.cn).}
  \thanks{G. Tang, H. Yang, Y. Song and G. Qian are with Intelligence Qubic
    (Beijing) Technology Co., Ltd, Beijing, China (e-mail: \{tgy, yanghang,
    songyu, grqian\}@iqubic.net).}}

\maketitle

\begin{abstract}
  As an emerging field, Automated Machine Learning (AutoML) aims to reduce or
  eliminate manual operations that require expertise in machine learning. In
  this paper, a graph-based architecture is employed to represent flexible
  combinations of ML models, which provides a large searching space compared to
  tree-based and stacking-based architectures. Based on this, an evolutionary
  algorithm is proposed to search for the best architecture, where the mutation
  and heredity operators are the key for architecture evolution. With Bayesian
  hyper-parameter optimization, the proposed approach can automate the workflow
  of machine learning. On the PMLB dataset, the proposed approach shows the
  state-of-the-art performance compared with TPOT, Autostacker, and
  auto-sklearn. Some of the optimized models are with complex structures which
  are difficult to obtain in manual design.

\end{abstract}

\section{Introduction}
\label{sec:introduction}

Various models have been thoroughly investigated by the machine learning (ML)
community. In theory, these models are general and applicable to both academia
and industry. However, it could be time-consuming to build a solution on a
specific ML task, even for a ML expert. To remedy this, Automated ML
(AutoML)~\cite{pmlr-v64-guyon_review_2016} emerges to minimize the design
complexity of a complete ML solution, which usually includes data preprocessing,
feature engineering, model selection and ensemble, fine-tuning of
hyperparameters, etc.

Research of AutoML starts from hyperparameter optimization, which includes
random search~\cite{bergstra_random_2012}, evolutionary
algorithm (EA)~\cite{friedrichs_evolutionary_2005}, and
Bayesian~\cite{brochu_tutorial_2010} approaches. Automatic feature
engineering~\cite{khurana_automating_2016,khurana_feature_2018} is another
important sub-field in AutoML.

In deep learning, the end-to-end scheme~\cite{krizhevsky_imagenet_2012} provides
a solution for automated learning to a certain degree. However, to train a deep
end-to-end neural network, it generally requires large scale labeled dataset,
which is unavailable in many practical problems. In addition,
Glasmachers has shown other limitations of the end-to-end scheme~\cite{glasmachers_limits_2017}.
At the mean while, neural architecture search approaches have
emerged~\cite{real_large-scale_2017,liu_progressive_2017} to automate neural
network design.

Though deep learning is very powerful, there is still room to apply traditional
ML. To achieve the goal of AutoML, Thornton et
al.~\cite{thornton_auto-weka_2013},
Thakur and Krohn-Grimberghe~\cite{thakur_autocompete_2015}, and
Feurer et al.~\cite{feurer_efficient_2015} proposed
solutions for automating the entire process for traditional ML
problems. These solutions are not powerful enough, because they use simple
combinations or single model selection. Employing EA
for architecture search, Tree-based Pipeline Optimization
Tool~\cite{olson_evaluation_2016}, or TPOT, uses a tree-based representation for
model combination; while Autostacker~\cite{chen_autostacker_2018} using the
stacking scheme.

According to a comparison between tree-based~\cite{cramer_representation_1985}
and graph-based~\cite{miller_cartesian_2011-1} representations for symbolic
regression~\cite{koza_genetic_1994}, a graph-based representation is much
flexible and general than the tree-based one. In addition, the graph-based
representation for computation are widely used in modern ML
frameworks including TensorFlow~\cite{abadi_tensorflow_2016}.

In this paper, we propose a novel AutoML solution called DarwinML based on the
EA with tournament selection~\cite{graff_evodag_2016}.
DarwinML employs the directed acyclic graph (DAG) for model combination.
Compared to existing AutoML methods such as pipeline, the proposed method is
with a flexible representation and is highly extensible.
In summary, the key contributions of this paper are as follows.
\begin{itemize}
\item The adjacency matrix of graph is analyzed and used to represent the
  architecture composed by a series of traditional ML models.
\item Several evolutionary operators have been defined and implemented to
  generate diverse architectures, which has not been thoroughly investigated in
  existing works.
\item Based on EA, an end-to-end automatic solution called ``DarwinML'' is
  proposed to search optimal composition of traditional ML models.
\end{itemize}

The rest of this paper is organized as follows. Section~\ref{sec:related-works}
reviews literature related to AutoML. Section~\ref{sec:arch-repr} introduces the
architectural representation for ML model composition.
Section~\ref{sec:evol-arch} presents the approach for evolutionary architecture
search and optimization. Section~\ref{sec:experiments} illustrates experimental
results on the PMLB~\cite{olson_pmlb_2017} dataset.
Section~\ref{sec:conclusions} concludes this work.

\section{Related Works}
\label{sec:related-works}

\subsection{Automatic Machine Learning}
\label{sec:autom-mach-learn}

The earlier AutoML models were born in the competition called ``ChaLearn AutoML
Challenge''~\cite{guyon_design_2015} in 2015. According to a test on 30
datasets, the two top ranking models of the challenge were designed by Intel's
team and Hutter's team, respectively. Intel's proprietary
solution~\cite{pmlr-v64-guyon_review_2016} is a tree-based method.
Auto-sklearn~\cite{feurer_efficient_2015}, the open-source solution from
Hutter's team, won the challenge. Hutter also co-developed
SMAC~\cite{hutter_sequential_2011} and
Auto-WEKA~\cite{thornton_auto-weka_2013,kotthoff_auto-weka_2017}. Auto-WEKA
treated AutoML as a Combined Algorithm Selection and Hyperparametric
optimization (CASH) problem. Based on open source package
WEKA~\cite{hall_weka_2009}, Auto-WEKA put traditional ML steps, including the
full range of classifiers and feature selectors, into one pipeline and uses
tree-based Bayesian optimization~\cite{hutter_sequential_2011} to search in the
combined space of algorithms and hyperparameters. Following Auto-WEKA,
auto-sklearn also solves the CASH problem with Bayesian hyperparameter
optimization (BHO). The improvements are the meta-learning initialization and
ensemble steps added before and after the step, respectively.
 
\subsection{Evolutionary Algorithms}
\label{sec:evol-algor}

EA has many variants in implementation such as
evolution programming, evolution strategy, genetic algorithm, etc. Encoding
problem to different representation that easier to evolve is an efficient way to
solution. Genetic Programming (GP), as a special type of EA, encodes computer
programs as a set of genes to evolve to the best. In recent years, inspired by
its success in generating structured
representations~\cite{gandomi_handbook_2015}, more
implementations~\cite{graff_semantic_2017,pawlak_semantic_2015} of GP has been
studied. One topic is using mathematical functions as coding units to search for
appropriate composite function structures. Such as graph-based representations
encoded by GP~\cite{graff_evodag_2016,ashlock_evolving_2015}, both approaches
use GP to evolve a better function to approximate target value.

In AutoML, EAs are used frequently for architecture search.
TPOT~\cite{olson_automating_2016}, a later AutoML approach which outperforms
previous systems, searches tree-based pipeline of models with GP. TPOT allows
dataset to be processed parallelly with multiple preprocessing methods. Such a
step can bring diverse features to the following steps. Furthermore,
Kordik et al.~\cite{kordik2018discovering} and de Sa et al.~\cite{de2017recipe}
not only use tree-based GP to ensemble ML methods, but also use a grammar to
enforce the generation of valid solution. The lager search space and detailed
constrains can make solution more like a successful method. Recently, Chen et
al. proposed Autostacker~\cite{chen_autostacker_2018} to combine ML classifiers
by stacking~\cite{ghorbani2001stacked} and search for the best model using EA.
This method largely extends the possibilities of the model combination.

\section{Architecture Representation}
\label{sec:arch-repr}

\subsection{Graph-based model combination}
\label{sec:graph-based-model}

A set of ML models, $\mathcal{M} = \{ f_{1}, \cdots, f_{M} \}$, is chosen to
construct the composite model, where $f_{m}$ denotes the hypothesis function of
the $m$-th model. A directed acyclic graph (DAG) is employed to denote the
architecture of a combination of models. In the DAG, edges and vertices denote
data flow and computational functions, respectively. This is the same to the
representation used in TensorFlow~\cite{abadi_tensorflow_2016}. One difference
is that the structure of DAG will be changed after evolutionary operations.

Let graph $\mathcal{G} = \{V, E\}$ be the DAG denoting the combined model, which
is composed of a set of vertices $V=\{v_{1}, \cdots, v_{K}\}$ and a set of edges
$E=\{ e_{ij} \}$. A vertex $v_{k}$ is associated with a computational function
from the model set $\mathcal{M}$. For simplicity, the model $f_{m_{k}}$ denotes
the function associated with the vertex $v_{k}$ in the rest of this paper. An
edge $e_{ij}$ implies that the output from vertex $v_{i}$ flows into the vertex
$v_{j}$.

Let $z_{j}$ denote the output of vertex $v_{j}$, and $z_{0}$ denote the input of
the composite model. Suppose vertex $v_{j}$ has $n$ input vertices,
$v_{i_{1}}, v_{i_{2}}, \cdots, v_{i_{n}}$, the function associated to vertex
$v_{j}$ is then computed as:
\begin{equation}
  \label{eq:vertex-function}
  z_{j} = f_{m_{j}}\bigg[U(z_{i_{1}}, z_{i_{2}}, \cdots, z_{i_{n}}) \bigg],
\end{equation}
where $U(\cdot)$ is a \emph{feature union} function. In the proposed approach,
vector concatenation is used as the feature union function. The output $z_{K}$
of the composite model can be computed by recursively applying
\eqref{eq:vertex-function}.

\subsection{Layer}
\label{sec:layer}

For the convenience of understanding and applying some evolutionary operations,
vertices are topologically sorted~\cite{kahn_topological_1962}. For simplicity,
the \emph{topological order} of a vertex is defined as its \emph{depth}.
Vertices with a same depth $d$ compose a \emph{layer} $L_{d}$:
\begin{equation}
  \label{eq:layer}
  L_{d} = \{ v_{k} | d_{k} = d, \forall v_{k} \in V \}.
\end{equation}
It should be noted that there are no edges between two vertices belong to a same
layer. Without loss of generality, vertices $v_{1}$ and $v_{K}$ are supposed to
be the input and the output of the graph, respectively. Accordingly, $d_{1}$ and
$d_{Z}$ are with the minimum and maximum depths.

The \emph{connection probability} of a particular edge $e_{ij}$ is useful for
some evolutionary operations, which is related to the depth of the two layers
containing vertices $v_{i}$ and $v_{j}$:
\begin{equation}
  \label{eq:connection-prob}
  p(e_{ij}) = p_{0} \exp\big( \gamma (d_{i} - d_{j} + 1) \big),
\end{equation}
where $\gamma$ is the decay factor, and $p_{0}$ is an initial connection
probability. 

\subsection{Layer blocks}
\label{sec:layer-blocks}

\begin{figure}
  \centering
  \includegraphics[width=3in]{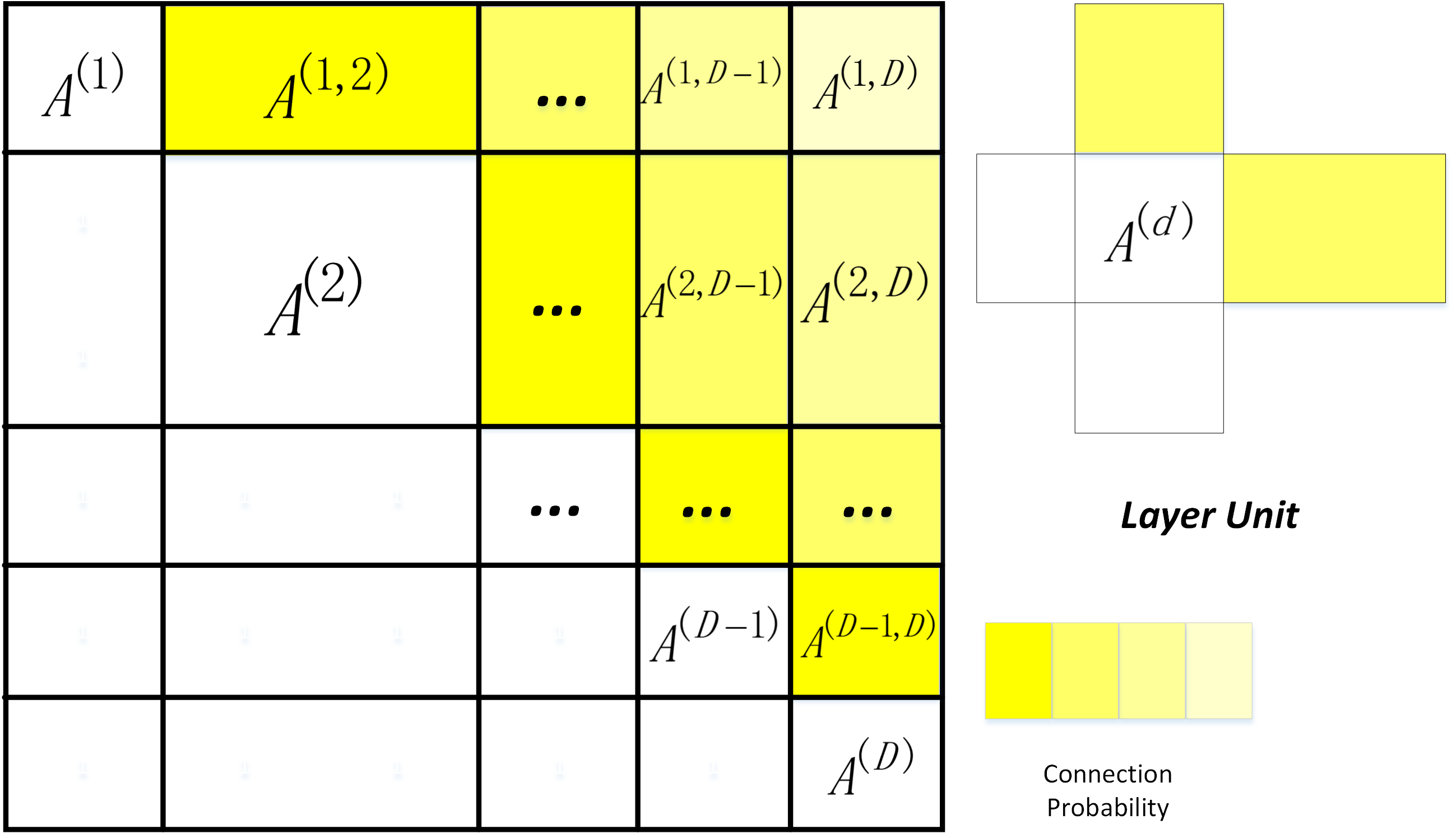}
  \caption{Layer blocks: The color corresponds to the connection probability
    between two layers $L_{d}$ and $L_{d'}$ calculated according
    to~\eqref{eq:block-prob}}
  \label{fig:layer-block}
\end{figure}

The \emph{adjacency matrix}, $A$, of a graph, $\mathcal{G}$, is a $K \times K$
matrix with elements given by:
\begin{equation}
  \label{eq:adj-mat}
  A_{ij} = \begin{cases} 1 & \text{if } e_{ij} \in E\\
    0 & \text{else} \end{cases},
\end{equation}
where one indicates an edge from vertex $v_{i}$ to vertex $v_{j}$. As the graph
is directed acyclic and topologically sorted, the adjacency matrix is always
upper triangular. As shown in Fig.~\ref{fig:layer-block}, the diagonal of the
adjacency matrix are square blocks full of zeros:
\begin{equation}
  \label{eq:zero-block}
  A_{ij}^{(d)} = 0 \qquad \forall v_{i}, v_{j} \in L_{d},
\end{equation}
where $A^{(d)}$ is a $|L_{d}| \times |L_{d}|$ block corresponds to the layer
$L_{d}$. The rest of the upper-triangle in the adjacency matrix can be split
into several blocks $A_{ij}^{(d,d')}$ according to the fragmenting of
layers, where $v_{i} \in L_{d}$ and $v_{j} \in L_{d'}$. The size of the
block $A^{(d,d')}$ is $|L_{d}| \times |L_{d'}|$.

The connection probability can be visualized by mapping onto the adjacency
matrix, as shown in Fig.~\ref{fig:layer-block}. According to
\eqref{eq:connection-prob}, each block $A^{(d,d')}$ has a same connection
probability which can be calculated as:
\begin{equation}
  \label{eq:block-prob}
  p \big( e_{ij} | v_{i} \in L_{d}, v_{j} \in L_{d'} \big)
  = p_{0} \exp\big( \gamma (d - d' + 1) \big).
\end{equation}
With above intuitive observation, changing layer $L_{d}$ should consider blocks
$A^{(d)}$, $A^{(d,\cdot)}$, and $A^{(\cdot, d)}$, which form a cross-shaped area
as shown in Fig.~\ref{fig:layer-block}.

\section{Evolution of Architectures}
\label{sec:evol-arch}

Our goal is to find the architecture that performs best on a given ML task. In
this paper, only classification tasks are considered and evaluated. Let
$X=\{x_{1}, x_{2}, \cdots, x_{N}\}$ and $Y=\{ y_{1}, y_{2}, \cdots, y_{N}\}$ be
the features and labels of a dataset, respectively. The optimization goal can
then be formally expressed as:
\begin{equation}
  \label{eq:goal}
  \mathcal{G}^{*}, \pmb{\Theta}^{*} = \arg \min_{\mathcal{G}, \pmb{\Theta}}
  \ell \big( \mathcal{G}, \pmb{\Theta}, X, Y \big)
  + \alpha c \big( \mathcal{G} \big),
\end{equation}
where $\mathcal{G}$ denotes the architecture represented by a DAG, and
$\pmb{\Theta}$ is the parameter endowed by the graph, $\ell(\cdot)$ is the loss
function of defined by the classification task, $c(\cdot)$ measures the
complexity of the graph, and $\alpha$ is a coefficient to trade-off between loss
and complexity. With the complexity term, the objective function prefers a
simple architecture. In implementation, the loss function includes a
regularization term and some cross validation strategies to avoid over-fitting,
and the complexity counts the number of vertices and edges in the graph. 

\subsection{Search Algorithm}
\label{sec:search-algorithm}

\begin{figure*}
  \centering
  \includegraphics[width=\linewidth]{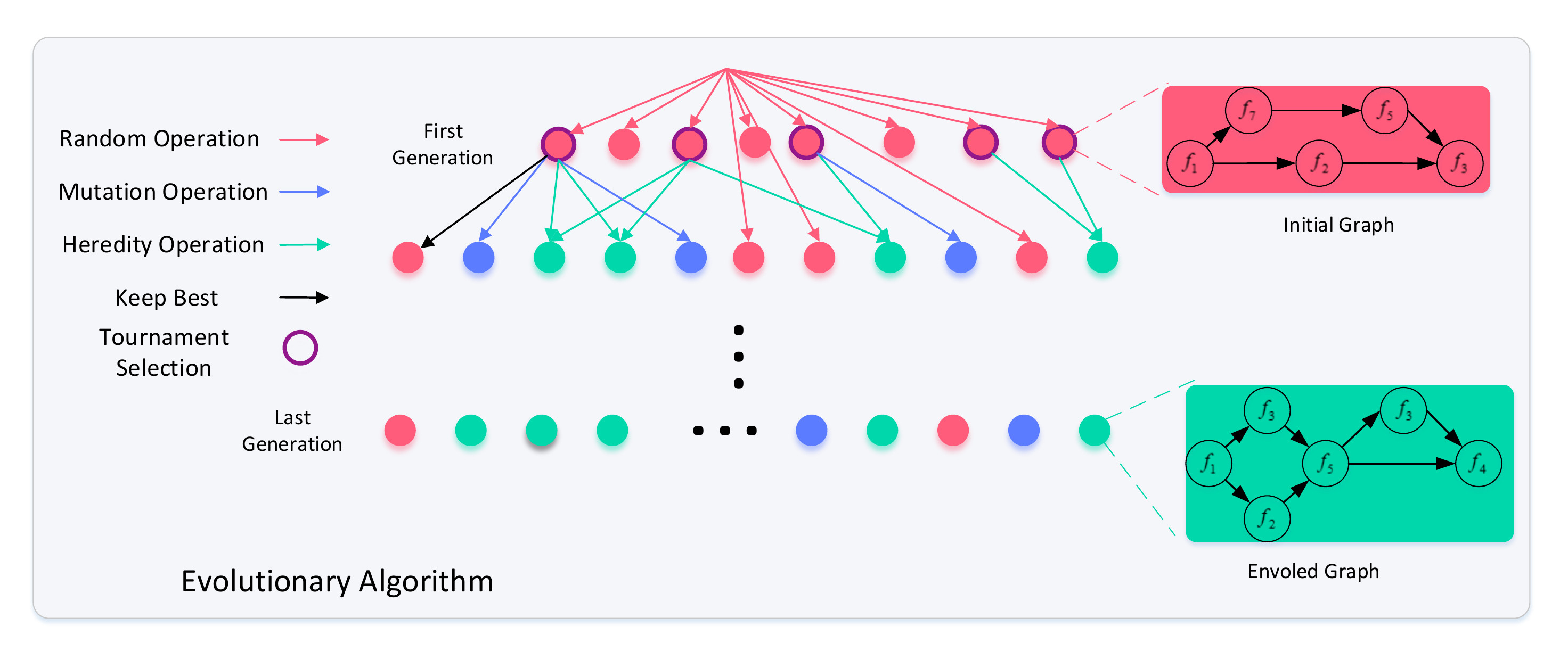}
  \caption{DarwinML framework. Evolutionary Algorithm(EA) using gene operations
    do searching job in the form of decoded matrices, and do training to get
    fitness in the form of encoded graph. After training, EA save best 15\%
    models to next generation. EA keep searching and training until the
    population is enough. The best model can be chosen in population by fitness
    as well as loss function in machine learning. }
  \label{fig:darwinml-ea}
\end{figure*}

The evolutionary search algorithm based on tournament
selection~\cite{goldberg_comparative_1992} is employed to explore for
architectures in the complex configuration space. The work flow is illustrated
in Fig.~\ref{fig:darwinml-ea}. Four operations, which are the random, mutation,
heredity, and keep best, constitute the main part of the algorithm. These
operations are designed to generate diverse architectures. Details of these
operations will be explained in following sub-sections. Following the convention
in EA, an architecture will also be called an \emph{individual} in the following.

In initialization, the first generation is build up by individuals according to
the convention in EA, generated by the random operation.

In the second generations, individuals evolve by applying all four
operations. Firstly, the keep best operator is applied to get top 15\% best individual set from first generation. Then, tournament selection\cite{goldberg_comparative_1992} picks the top
model from a randomly chosen sub-group in previous generation and keep best set. A number of top
models are selected by repeating random grouping and tournament selection.
Secondly, the mutation and heredity operations are applied to these promising
individuals, and new individuals are created for the current generation. 
Then,
the top 15\% best individuals are inherent back in the population to ensure
models with good performance are reserved. After that, individuals are generated
by the random operation to fulfill the current generation. Finally, fitness of
all individuals, except the ones inherent from keep best, are evaluated after a
training procedure.

In the subsequent generation, $i$-th generation applied by mutation and heredity  operation are produced from $(i-1)$-th generation and  top 15\% best individuals in $(i-2)$-th generation.

In every generation, the percentage of individuals generated from random,
heredity, and mutation operations are about 30\%, 40\%, and 30\%, respectively.
A higher heredity probability shows a better performance in tasks fit for
complex architectures. Individuals with invalid graph structures are simply
dropped no matter how the individual is generated. The rules will be explained in Implementation Details.  In addition, the search
algorithm prefers simple architectures according to the fitness in~\eqref{eq:goal}.

The evolution stops when a predefined duration or number of population has been
reached. The final output is the individual with the highest fitness.

\subsection{Evolutionary Operations}
\label{sec:evol-oper}

\subsubsection{Random Operation}
\label{sec:random-operation}

The random operation is designed to randomly sample individuals in the
configuration or searching space. The pseudo code of the random operation is
shown in Algorithm~\ref{algo:random}. In the algorithm, $K$ and $D$ are the
numbers of vertices and layers of the DAG to be generated as an individual. The
constant $C$ serves as an upper bound to control the size of DAG. The sizes of
layers, $l_{d}$, should follow a constraint that $\sum_{d}^{D} l_{d} = K$, where
$l_{1} = l_{D} = 1$ applies since the DAG is always implemented as
single-in-single-output. The predefined threshold $\rho$ controls the density of
edges.

\begin{algorithm}
  \caption{Random Operation}\label{algo:random}
  \begin{algorithmic}[1]
    \State $K \gets \mathtt{randint}$ \Comment{Number of vertices ($1<K<C$)}
    \State Randomly choose $K$ functions from the model set $\mathcal{M}$ and
    assign them to each vertex.

    \State $D \gets \mathtt{randint()}$
    \Comment{Number of layers ($D < K$).}

    \State $l_{2}, \cdots, l_{D-1} \gets \mathtt{randint()}$
    \Comment{Size of each layer.}
    \State $A \gets 0$ \Comment{Initialize the adjacency matrix.}
    \For{$1<d<d'<D$}
    \For{$v_{i}\in L_{d}, v_{j} \in L_{d'}$}
    \State $ p_{ij} \gets \mathtt{rand()}$ according to~\eqref{eq:block-prob}.
    \Comment{$0 \leq p_{ij} \leq 1$.}
    \State $A_{ij}^{(d,d')} \gets 1 \text{  If } p_{ij} > \rho$
    \Comment{Determine edge $e_{ij}$.}
    \EndFor
    \EndFor    
    \State Perform topological sort.
  \end{algorithmic}
\end{algorithm}

\subsubsection{Mutation Operations}
\label{sec:mutation-operations}

Three mutation operations are designed to provide flexible way to vary
individual architectures for the purpose of traversing to better individuals.

\begin{figure}
  \centering
  \includegraphics[width=0.75\linewidth]{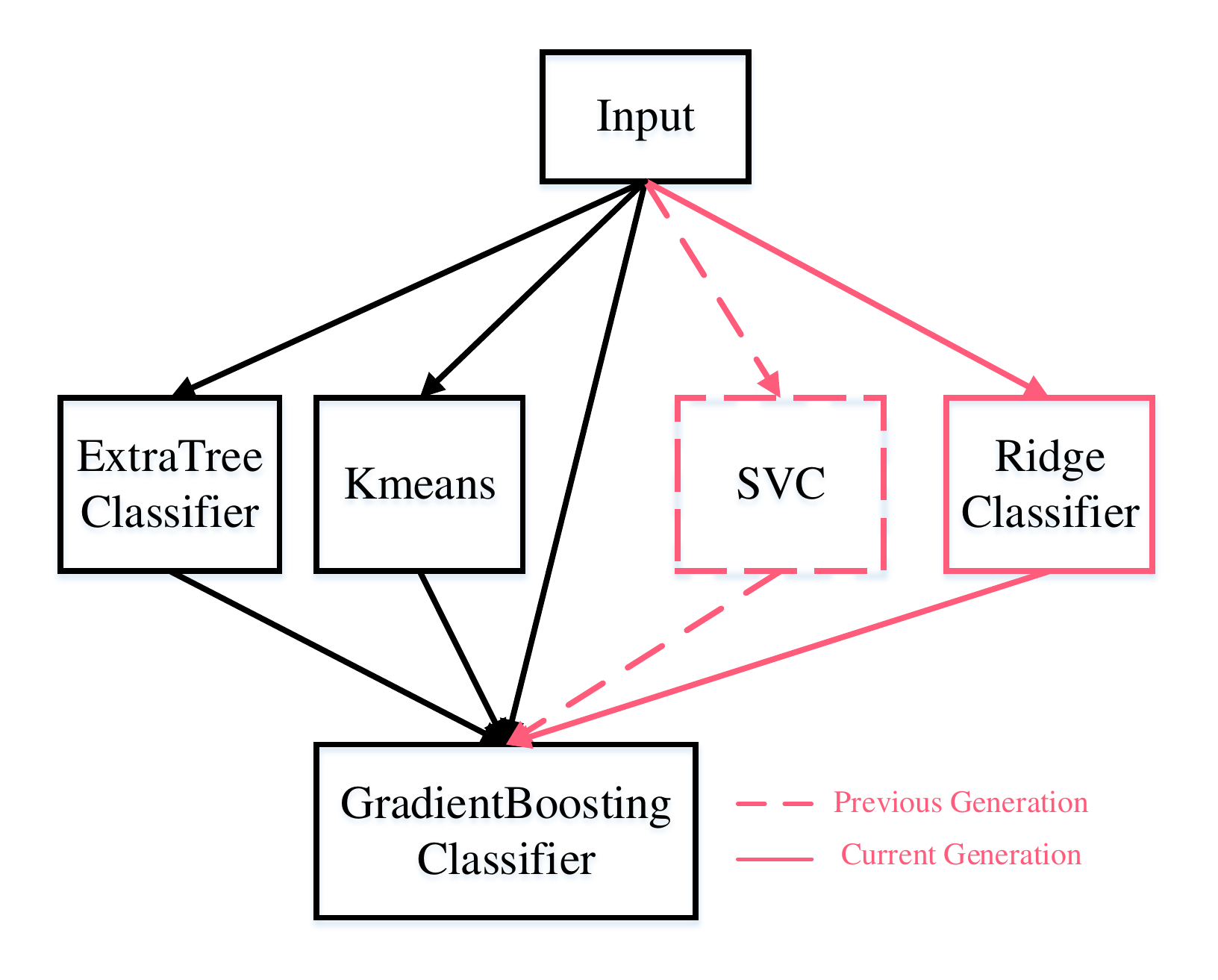}
  \caption{An example of vertex mutation.}
  \label{fig:vertex-mutation}
\end{figure}

\paragraph{Vertex mutation} This operation is designed to replace one vertex
with another ML model. Its implementation is as straightforward as shown by
Algorithm~\ref{algo:vertex-mutation}. A real example of this operation is shown
in Fig.~\ref{fig:vertex-mutation}, where the vertex model ``SVC'' is
replaced by a ``ridge classifier''.

\begin{algorithm}
  \caption{Vertex Mutation}\label{algo:vertex-mutation}
  \begin{algorithmic}[1]
    \State $k \gets \mathtt{randint()}$
    \Comment{Select vertex $v_{k}$ to change.}
    \State $m'_{k} \gets \mathtt{randint()}$
    \Comment{Choose a ML model from $\mathcal{M}$.}
    \State $f_{m_{k}} \gets f_{m'_{k}}$
    \Comment{Replace the model for vertex $v_{k}$.}
  \end{algorithmic}
\end{algorithm}

\paragraph{Edge mutation} This operation, as given by
Algorithm~\ref{algo:edge-mutation}, is defined to flip an edge connection in the
DAG.

\begin{algorithm}
  \caption{Edge Mutation}\label{algo:edge-mutation}
  \begin{algorithmic}[1]
    \State{$i,j \gets \mathtt{randint()}$}
    \Comment{Select two vertices $v_{i}$ and $v_{j}$.}
    \State{$A_{ij} \gets 1 - A_{ij}$}
    \Comment{Flip the connection of edge $e_{ij}$.}
    \State{Perform topological sort.}
  \end{algorithmic}
\end{algorithm}

\begin{figure}
  \centering
  \includegraphics[width=0.75\linewidth]{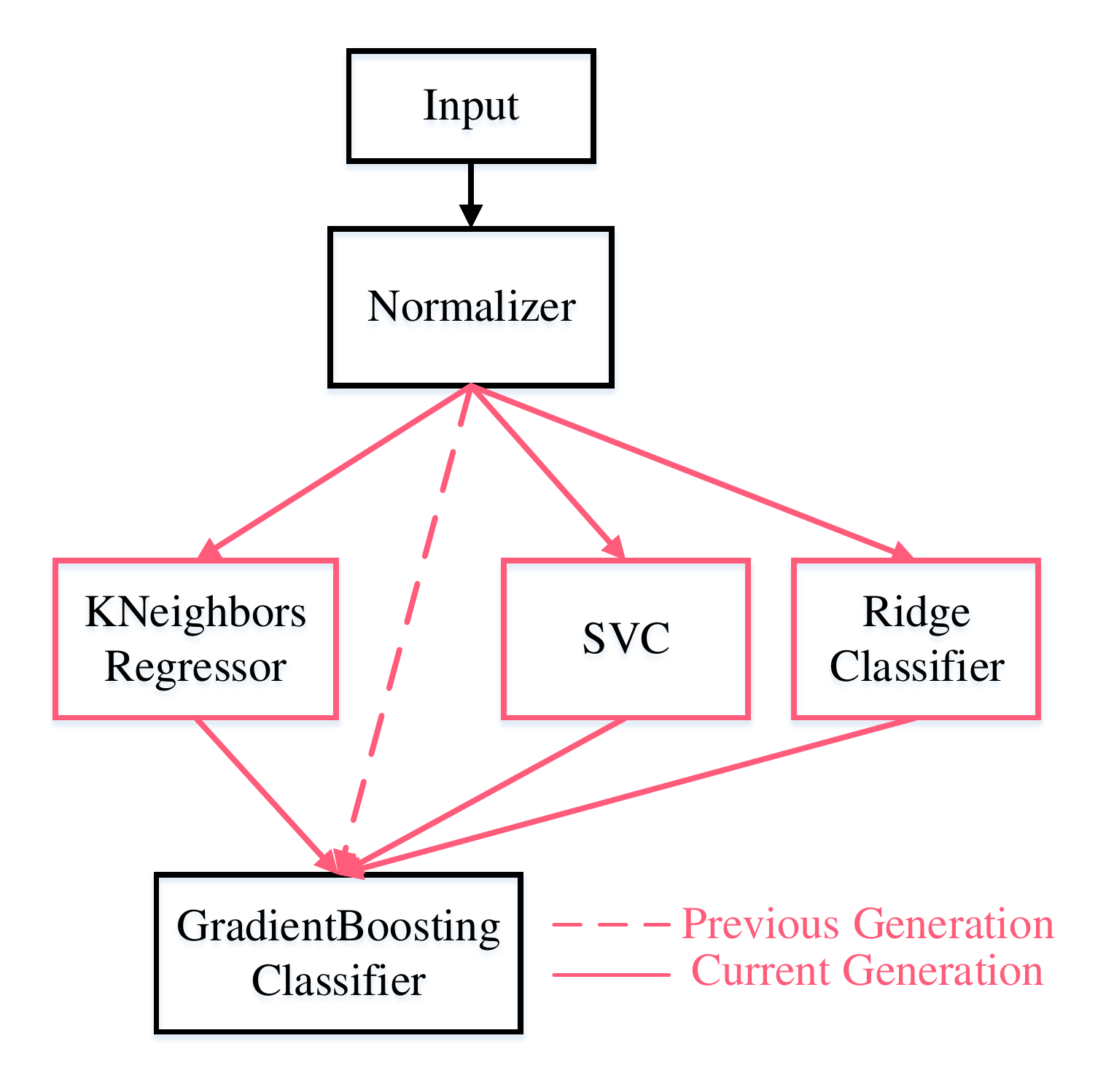}
  \caption{An example of layer mutation.}
  \label{fig:layer-mutation}
\end{figure}

\paragraph{Layer mutation} Local structures could affect the performance of the
architecture. Layers are naturally such local structures. Thus, the layer
mutation is introduced to change the DAG at a scale larger than vertex and edge.
The pseudo code is illustrated in Algorithm~\ref{algo:layer-mutation}. An
example is shown in Fig.~\ref{fig:layer-mutation}, where a layer composed by
three vertices, which are the ``KNeighbors Regressor'', ``SVC'', and ``Ridge
Classifier'', was inserted to the DAG. The dashed edge was automatically removed
after insertion. Please note that edges crossing layers are permitted but not
available in this example.

\begin{algorithm}
  \caption{Layer Mutation}\label{algo:layer-mutation}
  \begin{algorithmic}[1]
    \State{$d \gets \mathtt{randint()} $}
    \Comment{Choose layer $L_{d} (1<d<D)$.}
    \State{$p \gets \mathtt{rand()}$}
    \Comment{A real random number $0<p<1$.}
    \If{$p > 0.5$}
    \Comment{To remove the layer $L_{d}$.}
    \State{$A^{(d,d')}, A^{(d',d)} \gets 0$}
    \State{Remove $A^{(d)}, A^{(d,d')}, A^{(d',d)}$ from $A$.}
    \State{Remove vertices $v_{i} \in L_{d}$}
    \Else
    \Comment{To insert a layer $L_{d+1}$.}
    \State{$l_{d+1} \gets \mathtt{randint()}$}
    \State{Select $l_{d+1}$ models to create layer $L_{d+1}$.}
    \State{$A^{(d+1)}, A^{(d+1,\cdot)}, A^{(\cdot,d+1)} \gets 0$.}
    \State{Insert $A^{(d+1)}, A^{(d+1,\cdot)}, A^{(\cdot,d+1)}$ to $A$.}
    \For{$i,j$ in blocks $A^{(d+1,\cdot)}$ and $A^{(\cdot,d+1)}$}
    \State $ p_{ij} \gets \mathtt{rand()}$ according to~\eqref{eq:block-prob}.
    \State $A_{ij} \gets 1 \text{  If } p_{ij} > \rho$
    \EndFor
    \EndIf
    \State{Perform topological sort.}
  \end{algorithmic}
\end{algorithm}

\subsubsection{Heredity Operation}
\label{sec:heredity-operations}

Heredity operation is an enhancement on dealing with local structures. This
operation provides an opportunity for an individual to replace a bad layer with
a good one. In this sense, heredity plays a very important role to inherent and
broadcast the good local structures to the whole population. Different to
previous operations, heredity requires two good individuals to perform the
operation, as given by Algorithm~\ref{algo:heredity-operation}.

\begin{algorithm}
  \caption{Heredity Operation}\label{algo:heredity-operation}
  \begin{algorithmic}[1]
    \State{Choose two good graphs $\mathcal{G}$ and $\mathcal{G}'$ via
      tournament selection.}

    \State{Randomly choose two layers, $L_{d}$ and
      $L'_{d'}$, one from each graph.}

    \State{Remove the layer $L_{d}$ from graph $\mathcal{G}$.}
    \State{}
    \Comment{Pseudo codes like lines 4--6 in
      Algorithm~\ref{algo:layer-mutation}.}

    \State{Insert the layer $L'_{d'}$ into graph $\mathcal{G}$.}
    \State{}
    \Comment{Pseudo codes like lines 10--15 in
      Algorithm~\ref{algo:layer-mutation}.}    

    \State{Perform topological sort.}
  \end{algorithmic}
\end{algorithm}

\begin{figure}
  \centering
  \includegraphics[width=0.75\linewidth]{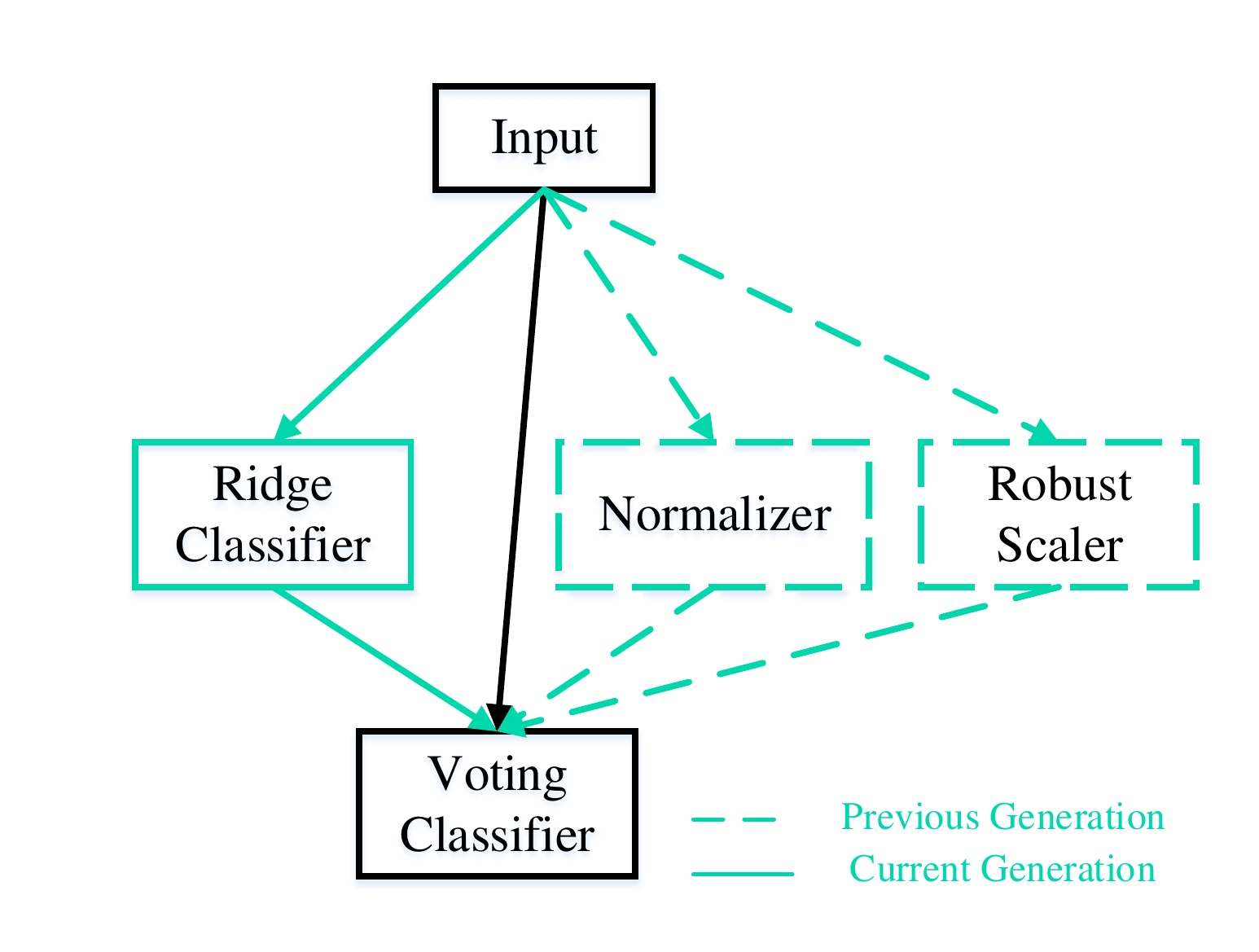}
  \caption{An example of heredity operation.}
  \label{fig:heredity-layer}
\end{figure}

One example is illustrated in Fig.~\ref{fig:heredity-layer}, the layer with one
vertex, ``Ridge Classifier'', has been inserted after the removal of the layer
containing two vertices, ``Normalizer'' and ``Robust Scaler''. With the heredity
operation, the balanced accuracy of the architecture increased to 0.778 from
0.723. The example shows the effectiveness of this operation.

\subsubsection{Keep Best}
\label{sec:keep-best}

This operation is used to keep the best individual in previous generation to
avoid losing the best one in following generations. In implementation, 15\%
individuals with top performances are treated as best ones and kept from one
generation to the next.

\subsection{Implementation Details}
\label{sec:implementation}

\subsubsection{Graph Validation}
\label{sec:validation-graph}

As the edges are generated by the evolutionary operations according to a random
distribution, there are possibilities that the generated individual is not a
valid DAG. So validation check should be performed after each operation,
and the invalid individual is dropped. There is a maximum retrial number for
these operations. So the percentages of individuals from each operation may
different to the configuration. One rule for validation check is that vertices
in a DAG should have both input and output edges except the first and last
vertex. Another rule can be applied to promote the ratio of individuals with
good performance, which is that classifier/regression vertices should not
directly connect to another classifier/regression vertex alone. Also, there are other constrains like maximum vertices number will be showed in experiment section.

\subsubsection{Model Mix}
\label{sec:model-mix}

After building a DAG with machine learning models, we have some problems while
the graph ensembles machine learning methods different from the task. 
If the task is classifier type and there is regression model in DAG, we
calculate root-mean-square error between regression prediction and classifier
label as regression loss function to train model. If unsupervised model in the graph, we do unsupervised learning on dataset and output a vector as feature to next vertex. For example, the output of k-means is a one-of-k coded vector, indicating the cluster index.
So the output of k-means can be used by its following vertex for model
combination (stacking).

\subsubsection{Hyperparameter Optimization}
\label{sec:hyperp-optim}

After searching models by the EA with tournament selection,
BHO~\cite{brochu_tutorial_2010} is applied to the top five individuals in the
final population. Due to the limitation of computing resources, hyperparameter
optimization is very expensive to apply to the whole population. The individuals
find by algorithm is competitive, and hyperparameter optimization could further
improve their performances.

\section{Experiments}
\label{sec:experiments}

To show the performance of DarwinML, its robustness and accuracy were evaluated.
Results were compared with random forest,
auto-sklearn~\cite{feurer_efficient_2015}, TPOT~\cite{olson_automating_2016},
and Autostacker~\cite{chen_autostacker_2018}. The capability of the model in
different search spaces and the influence of hyperparameter optimization are
demonstrated by ablation experiments. Interestingly, DarwinML found a variety of
interpretable solutions, which are also illustrated.

\subsection{Datasets of PMLB}
\label{sec:datasets-pmlb}

DarwinML was tested on the same datasets used in
Autostacker~\cite{chen_autostacker_2018}, where 15 datasets were selected from
PMLB~\cite{olson_pmlb_2017}. PMLB is a benchmark dataset that include hundreds
of public datasets which mainly resources from
OpenML~\cite{vanschoren_openml_2014} and UCI~\cite{dheeru_uci_2017}. These 15
datasets present various domains in PMLB that can test DarwinML's classification
performance on both binary and multiple-class tasks.
Table~\ref{tab:describe-pmlb} shows the detail of these datasets. For each
dataset, samples were shuffled and divided it into training (80\%) and testing
(20\%) sets. The training set uses cross validation.

\begin{table}
  \centering
	\begin{tabular}{lccc}
      \hline
      Datasets                  & \#Instances & \#Features & \#Classes \\
      \hline
      monk1                     & 556         & 6          & 2       \\
      parity5                   & 32          & 5          & 2       \\
      parity5+5                 & 1124        & 10         & 2       \\
      pima                      & 768         & 8          & 2       \\
      prnn\_crabs               & 200         & 7          & 2       \\
      allhypo                   & 3770        & 29         & 3       \\
      spect                     & 267         & 22         & 2       \\
      vehicle                   & 846         & 18         & 5       \\
      wine-recognition          & 178         & 13         & 4       \\
      breast-cancer             & 286         & 9          & 2       \\
      cars                      & 392         & 8          & 3       \\ 
      dis                       & 3772        & 29         & 2       \\ 
      Hill\_Valley\_with\_noise & 1212        & 100        & 2       \\ 
      ecoli                     & 327         & 7          & 8       \\ 
      heart-h                   & 294         & 13         & 2       \\ 
      \hline
	\end{tabular}
	\caption{Datasets used in experiments}
	\label{tab:describe-pmlb}
\end{table}

As listed in Table~\ref{tab:constrain-darwin}, two tests with different sizes of
population, which have 120 and 400 individuals to be evaluated in total, were
run to compare the performance with different search spaces. The corresponding
numbers of generation were set to 10 and 20, respectively. Results from the two
population sizes were named ``DML120'' and ``DML400'', respectively. In
addition, ranges of numbers of vertices and layers were set to constrain the
search space, as done in Kordik et al.~\cite{kordik2018discovering} and de Sa et
al.~\cite{de2017recipe}. The number of retrials is the upper limit of the
failures in trying to apply evolutionary operators to a graph. If this limit is
reached, the individual will be dropped according to
sub-section~\ref{sec:validation-graph}. The max training time is the upper limit
of the training time can be used for one graph. If time is out, graph will be
dropped as a failure graph. The parameters $p_{0}$ and $\gamma$
in~\eqref{eq:connection-prob} are set to 0.3 and 1, respectively. The parameter
$C$ in random operation is set to 10.

Results named ``DML400+BHO'' are obtained by applying
BHO~\cite{brochu_tutorial_2010} to the top five best graphs in the population
with 400 individuals. With Bayesian optimization, we tuned up to 40 parameter
sets. According to Autostacker~\cite{chen_autostacker_2018} and
TPOT~\cite{olson_automating_2016}, the balanced
accuracy~\cite{velez_balanced_2007} is employed to measure the performance for a
fair comparison. On each dataset, DarwinML were repeated 10 times with random
initialization, the mean and variances of the balanced accuracy are calculated.
Results of other AutoML methods are collected from the paper of
Autostacker~\cite{chen_autostacker_2018}, which is produced with a same setting.

\begin{table}
  \centering
  \begin{tabular}{lc}
    \hline
    Parameters              & Configuration         \\ \hline
    \#populations             & 120/400               \\ 
    \#generations             & 10/20                 \\
    \#vertices              & 2--12                \\
    \#layers                & 2--6                 \\
    \#retrials              & 100                   \\ 
    max training time       & 3600 secs             \\
    \hline
  \end{tabular}
  \caption{Constrain configuration used in experiments}
  \label{tab:constrain-darwin}
\end{table}

\begin{table*}
	\centering
	\resizebox{\textwidth}{30mm}{
	\begin{tabular}{lccccccc}
		\hline
		Datasets & RandomForest & auto-sklearn & TPOT & Autostacker & DML120 & DML400 & DML400+BHO \\
		\hline
monk1 & 0.98$\pm $0.009 & \textbf{1$\pm $0}   & \textbf{1$\pm $0}   & \textbf{1$\pm $0}   & \textbf{1$\pm $0}   & \textbf{1$\pm $0}   & \textbf{1$\pm $0} \\

parity5 & 0.02$\pm $0.053 & 0.87$\pm $0.209 & 0.81$\pm $0.21 & 0.94$\pm $0.138 & \textbf{1$\pm $0}   & \textbf{1$\pm $0}   & \textbf{1$\pm $0} \\

parity5+5 & 0.60$\pm $0.050 & \textbf{1$\pm $0}   & \textbf{1$\pm $0}   & \textbf{1$\pm $0}   & 0.88$\pm $0.044 & 0.93$\pm $0.030 & \textbf{1$\pm $0} \\

pima  & 0.73$\pm $0.033 & 0.72$\pm $0.040 & 0.73$\pm $0.05 & 0.74$\pm $0.023 & 0.74$\pm $0.009 & 0.77$\pm $0.006 & \textbf{0.79$\pm $0.009} \\

prnn\_crabs & 0.95$\pm $0.027 & 0.99$\pm $0.019 & 1$\pm $0.008 & \textbf{1$\pm $0}   & \textbf{1$\pm $0}   & \textbf{1$\pm $0}   & \textbf{1$\pm $0} \\

allhypo & 0.79$\pm $0.021 & 0.89$\pm $0.029 & 0.95$\pm $0.046 & 0.94$\pm $0.026 & 0.86$\pm $0.025 & 0.87$\pm $0.015 & \textbf{0.97$\pm $0.003} \\

spect & 0.68$\pm $0.068 & 0.71$\pm $0.046 & 0.81$\pm $0.031 & 0.82$\pm $0.04 & 0.83$\pm $0.024 & 0.85$\pm $0.013 & \textbf{0.86$\pm $0.010} \\

vehicle & 0.83$\pm $0.021 & \textbf{0.90$\pm $0.017} & 0.82$\pm $0.039 & 0.89$\pm $0.044 & 0.84$\pm $0.012 & 0.86$\pm $0.007 & 0.85$\pm $0.005 \\

wine-recognition & 0.99$\pm $0.015 & 0.97$\pm $0.021 & 0.98$\pm $0.018 & 0.99$\pm $0.012 & \textbf{1$\pm $0}   & \textbf{1$\pm $0}   & \textbf{1$\pm $0} \\

breast-cancer & 0.59$\pm $0.058 & 0.59$\pm $0.059 & 0.67$\pm $0.090 & 0.66$\pm $0.080 & 0.69$\pm $0.015 & 0.71$\pm $0.012 & \textbf{0.77$\pm $0.017} \\

cars  & 0.91$\pm $0.034 & 0.97$\pm $0.013 & 0.96$\pm $0.036 & 0.98$\pm $0.014 & 0.93$\pm $0.022 & 0.96$\pm $0.018 & \textbf{0.99$\pm $0.007} \\

dis   & 0.55$\pm $0.042 & 0.68$\pm $0.069 & 0.76$\pm $0.061 & 0.79$\pm $0.046 & 0.79$\pm $0.033 & 0.82$\pm $0.022 & \textbf{0.90$\pm $0.003} \\

Hill\_Valley\_with\_noise & 0.56$\pm $0.027 & \textbf{1$\pm $0.003} & 0.96$\pm $0.043 & 0.98$\pm $0.015 & 0.98$\pm $0.013 & 0.97$\pm $0.010 & 0.97$\pm $0.013 \\

ecoli & 0.91$\pm $0.030 & 0.89$\pm $0.062 & 0.86$\pm $0.043 & 0.92$\pm $0.029 & 0.82$\pm $0.030 & 0.85$\pm $0.016 & \textbf{0.95$\pm $0.013} \\

heart-h & 0.79$\pm $0.036 & 0.79$\pm $0.042 & 0.81$\pm $0.047 & 0.83$\pm $0.022 & 0.86$\pm $0.013 & 0.85$\pm $0.018 & \textbf{0.86$\pm $0.008} \\

		\hline
	\end{tabular}}
	\caption{Test Accuracy Comparison. Results on same 15 PMLB Datasets in
      Autostacker.}
	\label{tab:db-pmlb}
\end{table*}

\subsection{Comparison}
\label{sec:Comparison}

Results were listed in Table~\ref{tab:db-pmlb}. Random forest is chosen to be
the baseline.
Being EA based AutoML models, TPOT~\cite{olson_automating_2016} and
Autostacker~\cite{chen_autostacker_2018} evolved for 100 and 3 generations,
respectively. Both models include hyperparameter optimization, which improves
the performance dramatically. In testing DarwinML, hyperparameter optimization
was separated for a detailed comparison. DML400 achieves better performance when
compared with random forest on 14 datasets. Its accuracy is comparable or
superior to TPOT on 11 datasets, and is comparable or superior to Autostacker
and auto-sklearn 9 datasets. Although other models except random forest include
hyperparameter optimization, DML400 is better than them because DarwinML's
inherent mechanism provides flexible model combination which enlarges the
architectural search space. Taking BHO as a post-processing step on DML400,
DarwinML far exceeds other models in mean balanced accuracy on most datasets. 

\begin{figure}
  \centering
  \includegraphics[width=\linewidth]{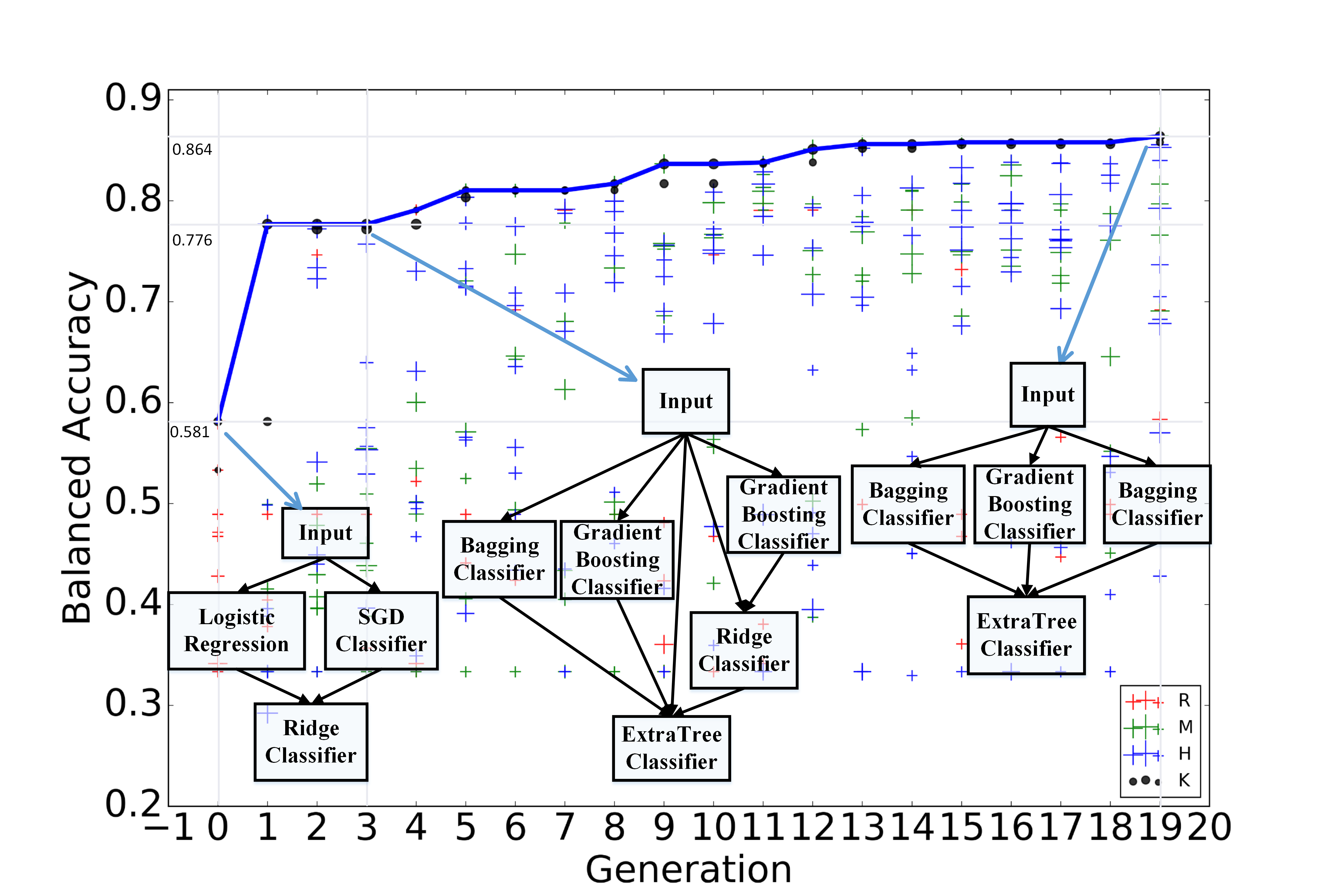}
  \caption{The scatter plot shows the performance evolution when architectures
    is searched on the ``{allhypo}'' dataset in PMLB~\cite{olson_pmlb_2017} with
    DarwinML. Three architectures are selected to show the flexibility of
    proposed evolutionary operations. Each point represents balanced accuracy
    evaluated by an individual graph. The point in the curve means the best
    individual in each generation. }
  \label{fig:evolution}
\end{figure}

Fig.~\ref{fig:evolution} shows the scatter plot of accuracy of each individual
on the dataset allhypo, which is one run of DML400. In the plot, the color of
each point represents how it is generated.
Some individuals run out of the training time and were dropped. Their loss were
set to infinity and were excluded from drawing on the plot. In the last few
generations of the evolution, individuals generated by random operation appear
less and less, while individuals from heredity and mutation operations are more
important for improving the performance. According to Fig.~\ref{fig:evolution},
the accuracy of best individuals increased continuously from 0.581 to 0.864. In
addition, it can be observed in each generation that performances of individuals
from heredity and mutation are generally with a performance better than the ones
sampled randomly. Some specific examples of applying these evolutionary
operations have been observed.
Fig.~\ref{fig:vertex-mutation} is a concrete example on dataset
``breast-cancer'' where the accuracy increased from 0.537 to 0.728 after
applying the vertex mutation. Fig.~\ref{fig:layer-mutation} shows a layer
mutation applied on ``Hill\_Vally\_with\_noise'' which increased the accuracy
from 0.971 to 1.000. Accuracy increment from 0.723 to 0.778 was observed in
experiments on pima after applying the heredity operation, as shown in
Fig.~\ref{fig:heredity-layer}. These observations demonstrate that DarwinML
provides rational and efficient operators to evolve individuals better.

\begin{figure}
\begin{subfigure}{0.475\linewidth}
	\includegraphics[width=\linewidth]{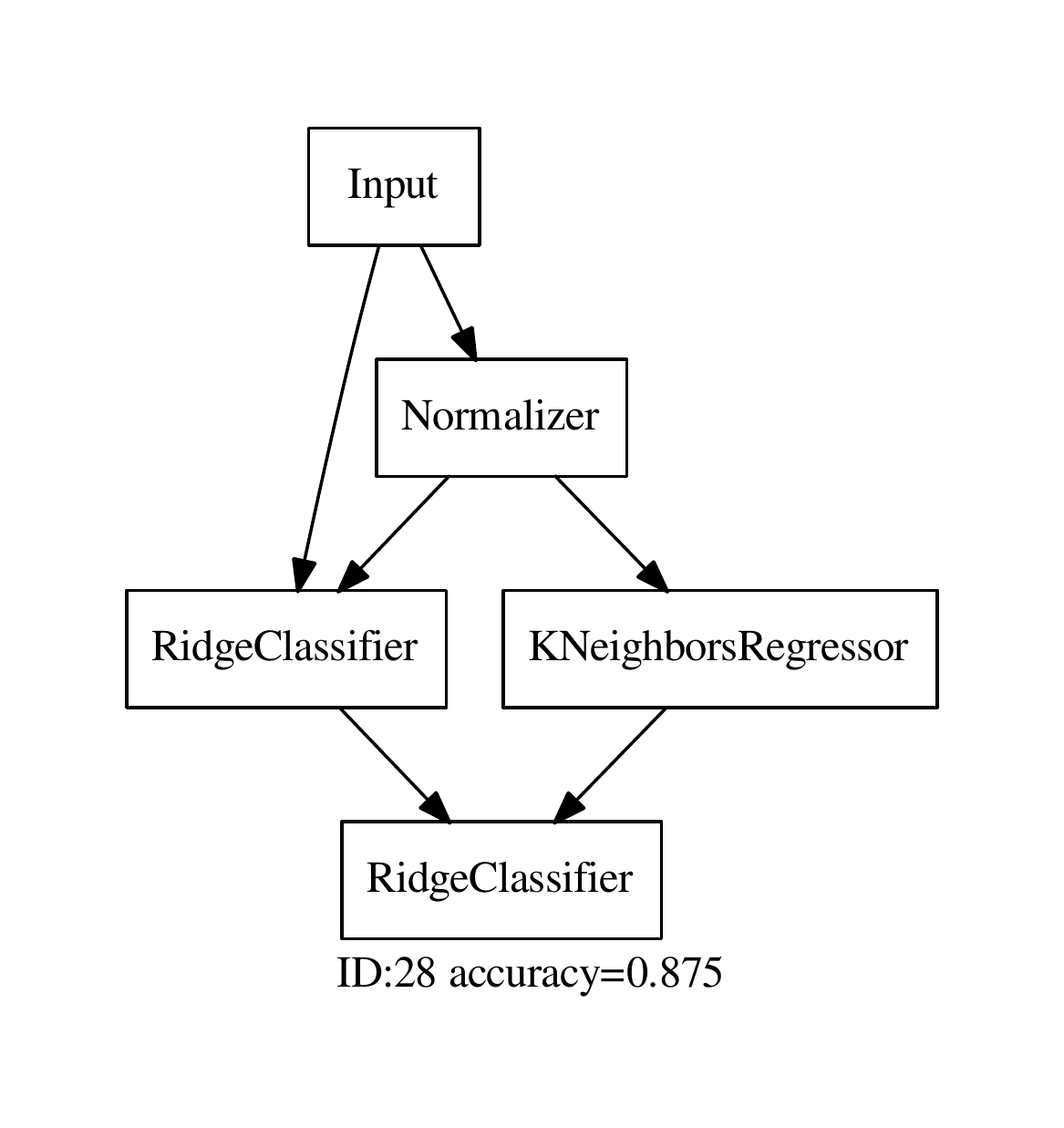}
	\caption{Second Generation}\label{fig:heart-h120:a}
\end{subfigure}
\begin{subfigure}{0.475\linewidth}
	\includegraphics[width=\linewidth]{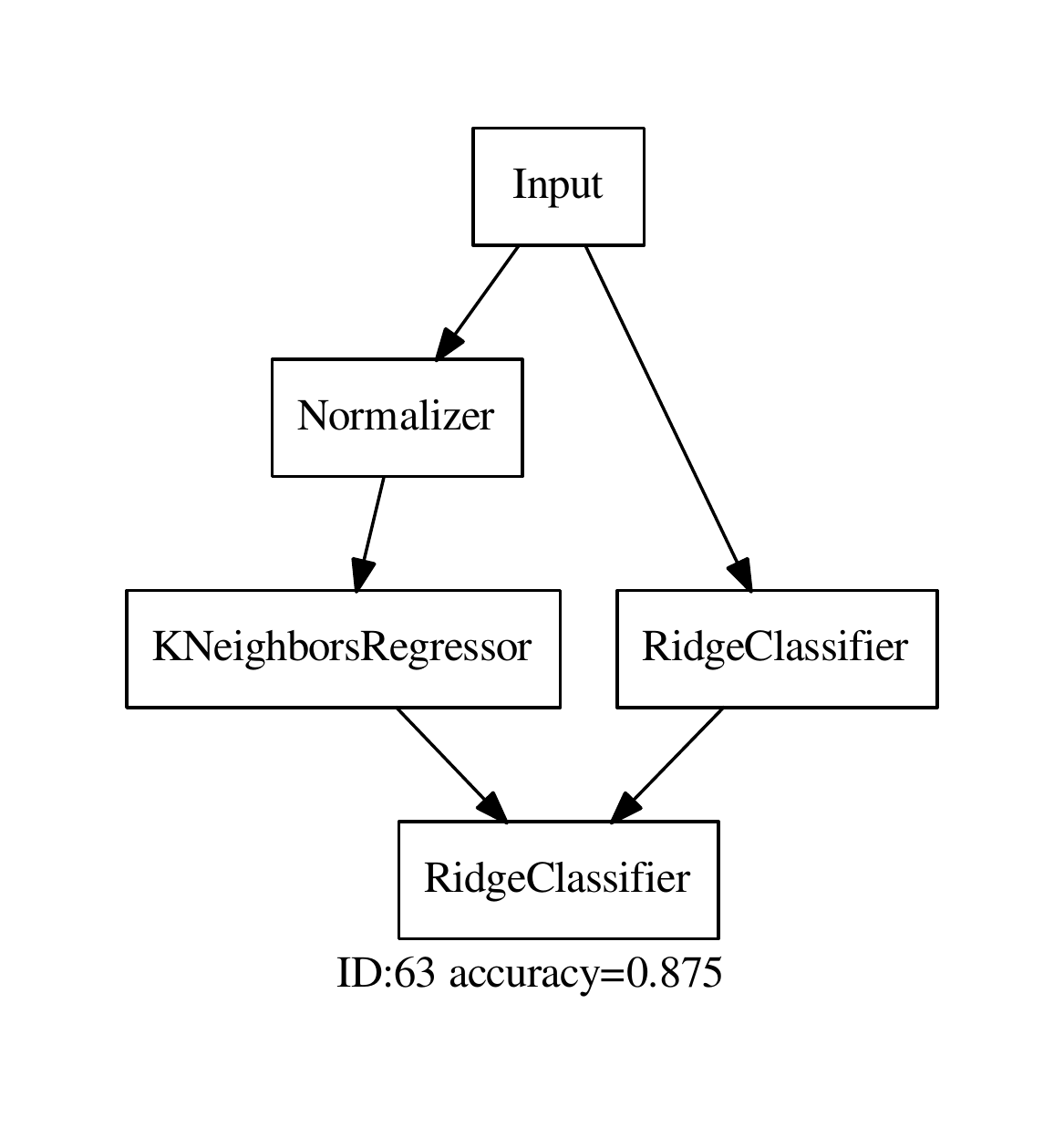}
	\caption{Third Generation}\label{fig:heart-h120:b}
\end{subfigure}
\caption{Two individual graphs on heart-h dataset generated by DarwinML120.}
\label{fig:heart-h120}
\end{figure}

In Fig.~\ref{fig:heart-h120}, we present best solutions on heart-h datasets
searched by DML120. Surprisingly, we observed that there are more than one
individuals achieved the best balanced accuracy, 0.875. The individual shown in
Fig.~\ref{fig:heart-h120:b} has one edge less than that in
Fig.~\ref{fig:heart-h120:a}. According to~\eqref{eq:goal}, DarwinML prefers for
a simpler structure if models have a nearly same loss $\ell(\cdot)$. So DarwinML
can find best architectures with respect to~\eqref{eq:goal}.
In addition, we also observed optimized architectures with complex structures in
experiments, which are difficult to obtain in manual design.

\section{Conclusions}
\label{sec:conclusions}

In this paper, an evolutionary algorithm is proposed to search for the best
architecture composed of traditional machine learning models with a graph-based
representation. Based on the representation, the random, mutation, and heredity
operators are defined and implemented. Evolutionary algorithm is then employed
to optimize the architecture. Diversity makes the success of EAs. The
evolutionary operations proposed in the paper enables the search of diverse
architectures. With Bayesian hyperparameter optimization applied to the best results of the proposed evolutionary method, the proposed approach
demonstrates the state-of-the-art performance on the PMLB dataset compared to
TPOT, auto-stacker, and auto-sklearn.

Though the proposed approach is implemented and tested on the set of traditional
machine learning models, there are no inherent limitations that it applies only
on these models. In the future, we plan to generalize the approach to neural
architecture search by extending the ML models with building blocks of neural
networks to get better performance on large scale datasets.

\bibliography{refs}

\begin{thebibliography}{10}
\providecommand{\url}[1]{#1}
\csname url@samestyle\endcsname
\providecommand{\newblock}{\relax}
\providecommand{\bibinfo}[2]{#2}
\providecommand{\BIBentrySTDinterwordspacing}{\spaceskip=0pt\relax}
\providecommand{\BIBentryALTinterwordstretchfactor}{4}
\providecommand{\BIBentryALTinterwordspacing}{\spaceskip=\fontdimen2\font plus
\BIBentryALTinterwordstretchfactor\fontdimen3\font minus
  \fontdimen4\font\relax}
\providecommand{\BIBforeignlanguage}[2]{{%
\expandafter\ifx\csname l@#1\endcsname\relax
\typeout{** WARNING: IEEEtran.bst: No hyphenation pattern has been}%
\typeout{** loaded for the language `#1'. Using the pattern for}%
\typeout{** the default language instead.}%
\else
\language=\csname l@#1\endcsname
\fi
#2}}
\providecommand{\BIBdecl}{\relax}
\BIBdecl

\bibitem{pmlr-v64-guyon_review_2016}
I.~Guyon, I.~Chaabane, H.~J. Escalante, S.~Escalera, D.~Jajetic, J.~R. Lloyd,
  N.~Maci\`a, B.~Ray, L.~Romaszko, M.~Sebag, A.~Statnikov, S.~Treguer, and
  E.~Viegas, ``A brief {{Review}} of the {{ChaLearn AutoML Challenge}}:
  {{Any}}-time {{Any}}-dataset {{Learning}} without {{Human Intervention}},''
  in \emph{Proceedings of the {{Workshop}} on {{Automatic Machine Learning}}},
  ser. PMLR, F.~Hutter, L.~Kotthoff, and J.~Vanschoren, Eds., vol.~64, New
  York, New York, USA, 2016, pp. 21--30.

\bibitem{bergstra_random_2012}
J.~Bergstra and Y.~Bengio, ``Random {{Search}} for {{Hyper}}-{{Parameter
  Optimization}},'' \emph{J. Mach. Learn. Res.}, vol.~13, no. Feb, pp.
  281--305, 2012.

\bibitem{friedrichs_evolutionary_2005}
F.~Friedrichs and C.~Igel, ``Evolutionary tuning of multiple {{SVM}}
  parameters,'' \emph{Neurocomputing}, vol.~64, pp. 107--117, Mar. 2005.

\bibitem{brochu_tutorial_2010}
E.~Brochu, V.~M. Cora, and N.~{de Freitas}, ``A {{Tutorial}} on {{Bayesian
  Optimization}} of {{Expensive Cost Functions}}, with {{Application}} to
  {{Active User Modeling}} and {{Hierarchical Reinforcement Learning}},''
  \emph{arXiv:1012.2599}, Dec. 2010.

\bibitem{khurana_automating_2016}
U.~Khurana, F.~Nargesian, H.~Samulowitz, {Elias Khalil}, and {Deepak Turaga},
  ``Automating {{Feature Engineering}},'' in \emph{{{AI4DS}}}, 2016.

\bibitem{khurana_feature_2018}
U.~Khurana, H.~Samulowitz, and D.~Turaga, ``\BIBforeignlanguage{en}{Feature
  {{Engineering}} for {{Predictive Modeling Using Reinforcement Learning}}},''
  in \emph{\BIBforeignlanguage{en}{Thirty-{{Second AAAI Conference}} on
  {{Artificial Intelligence}}}}, Apr. 2018.

\bibitem{krizhevsky_imagenet_2012}
A.~Krizhevsky, I.~Sutskever, and G.~E. Hinton, ``{{ImageNet Classification}}
  with {{Deep Convolutional Neural Networks}},'' in \emph{Proc. Adv. Neural
  Inf. Process. Syst. ({NIPS})}, F.~Pereira, C.~J.~C. Burges, L.~Bottou, and
  K.~Q. Weinberger, Eds.\hskip 1em plus 0.5em minus 0.4em\relax {Curran
  Associates, Inc.}, 2012, pp. 1097--1105.

\bibitem{glasmachers_limits_2017}
T.~Glasmachers, ``\BIBforeignlanguage{en}{Limits of {{End}}-to-{{End
  Learning}}},'' in \emph{\BIBforeignlanguage{en}{Asian {{Conference}} on
  {{Machine Learning}}}}, Nov. 2017, pp. 17--32.

\bibitem{real_large-scale_2017}
E.~Real, S.~Moore, A.~Selle, S.~Saxena, Y.~L. Suematsu, J.~Tan, Q.~V. Le, and
  A.~Kurakin, ``\BIBforeignlanguage{en}{Large-{{Scale Evolution}} of {{Image
  Classifiers}}},'' in \emph{\BIBforeignlanguage{en}{Proc. Int. Conf. Mach.
  Learn. ({ICML})}}, Jul. 2017, pp. 2902--2911.

\bibitem{liu_progressive_2017}
C.~Liu, B.~Zoph, M.~Neumann, J.~Shlens, W.~Hua, L.-J. Li, L.~{Fei-Fei},
  A.~Yuille, J.~Huang, and K.~Murphy, ``Progressive {{Neural Architecture
  Search}},'' \emph{arXiv:1712.00559}, Dec. 2017.

\bibitem{thornton_auto-weka_2013}
C.~Thornton, F.~Hutter, H.~H. Hoos, and K.~{Leyton-Brown}, ``Auto-{{WEKA}}:
  {{Combined Selection}} and {{Hyperparameter Optimization}} of
  {{Classification Algorithms}},'' in \emph{Proceedings of the 19th {{ACM
  SIGKDD International Conference}} on {{Knowledge Discovery}} and {{Data
  Mining}}}, ser. KDD '13, New York, NY, USA, 2013, pp. 847--855.

\bibitem{thakur_autocompete_2015}
A.~Thakur and A.~{Krohn-Grimberghe}, ``{{AutoCompete}}: {{A Framework}} for
  {{Machine Learning Competition}},'' \emph{arXiv:1507.02188}, Jul. 2015.

\bibitem{feurer_efficient_2015}
M.~Feurer, A.~Klein, K.~Eggensperger, J.~Springenberg, M.~Blum, and F.~Hutter,
  ``Efficient and robust automated machine learning,'' in \emph{Proc. Adv.
  Neural Inf. Process. Syst. ({NIPS})}, C.~Cortes, N.~D. Lawrence, D.~D. Lee,
  M.~Sugiyama, and R.~Garnett, Eds.\hskip 1em plus 0.5em minus 0.4em\relax
  {Curran Associates, Inc.}, 2015, pp. 2962--2970.

\bibitem{olson_evaluation_2016}
R.~S. Olson, N.~Bartley, R.~J. Urbanowicz, and J.~H. Moore, ``Evaluation of a
  {{Tree}}-based {{Pipeline Optimization Tool}} for {{Automating Data
  Science}},'' in \emph{Proceedings of the {{Genetic}} and {{Evolutionary
  Computation Conference}} 2016}, ser. GECCO'16.\hskip 1em plus 0.5em minus
  0.4em\relax New York, NY, USA: {ACM}, 2016, pp. 485--492.

\bibitem{chen_autostacker_2018}
B.~Chen, H.~Wu, W.~Mo, I.~Chattopadhyay, and H.~Lipson, ``Autostacker: {{A
  Compositional Evolutionary Learning System}},'' in \emph{Proceedings of the
  {{Genetic}} and {{Evolutionary Computation Conference}}}, ser.
  GECCO'18.\hskip 1em plus 0.5em minus 0.4em\relax New York, NY, USA: {GECCO},
  2018, pp. 402--409.

\bibitem{cramer_representation_1985}
N.~L. Cramer, ``A {{Representation}} for the {{Adaptive Generation}} of
  {{Simple Sequential Programs}},'' in \emph{Proceedings of the 1st
  {{International Conference}} on {{Genetic Algorithms}}}.\hskip 1em plus 0.5em
  minus 0.4em\relax Hillsdale, NJ, USA: {L. Erlbaum Associates Inc.}, 1985, pp.
  183--187.

\bibitem{miller_cartesian_2011-1}
J.~F. Miller, ``Cartesian {{Genetic Programming}},'' in \emph{Cartesian
  {{Genetic Programming}}}, ser. Natural Computing Series.\hskip 1em plus 0.5em
  minus 0.4em\relax {Springer-Verlag Berlin Heidelberg}, 2011, pp. 17--34.

\bibitem{koza_genetic_1994}
J.~R. Koza, \emph{Genetic {{Programming II}}: {{Automatic Discovery}} of
  {{Reusable Programs}}}.\hskip 1em plus 0.5em minus 0.4em\relax Cambridge, MA,
  USA: {MIT Press}, 1994.

\bibitem{abadi_tensorflow_2016}
M.~Abadi, P.~Barham, J.~Chen, Z.~Chen, A.~Davis, J.~Dean, M.~Devin,
  S.~Ghemawat, G.~Irving, M.~Isard, M.~Kudlur, J.~Levenberg, R.~Monga,
  S.~Moore, D.~G. Murray, B.~Steiner, P.~Tucker, V.~Vasudevan, P.~Warden,
  M.~Wicke, Y.~Yu, and X.~Zheng, ``{{TensorFlow}}: {{A System}} for
  {{Large}}-{{Scale Machine Learning}},'' in \emph{12th {{USENIX Symposium}} on
  {{Operating Systems Design}} and {{Implementation}} ({{OSDI}} 16)}.\hskip 1em
  plus 0.5em minus 0.4em\relax Savannah, GA: {USENIX Association}, 2016, pp.
  265--283.

\bibitem{graff_evodag_2016}
M.~Graff, E.~S. Tellez, S.~{Miranda-Jim\'enez}, and H.~J. Escalante,
  ``{{EvoDAG}}: {{A}} semantic {{Genetic Programming Python}} library,'' in
  \emph{2016 {{IEEE International Autumn Meeting}} on {{Power}},
  {{Electronics}} and {{Computing}} ({{ROPEC}})}, Nov. 2016, pp. 1--6.

\bibitem{olson_pmlb_2017}
R.~S. Olson, W.~La~Cava, P.~Orzechowski, R.~J. Urbanowicz, and J.~H. Moore,
  ``{{PMLB}}: A large benchmark suite for machine learning evaluation and
  comparison,'' \emph{BioData Mining}, vol.~10, p.~36, Dec. 2017.

\bibitem{guyon_design_2015}
I.~Guyon, K.~Bennett, G.~Cawley, H.~J. Escalante, S.~Escalera, T.~K. Ho,
  N.~Maci\`a, B.~Ray, M.~Saeed, A.~Statnikov, and E.~Viegas, ``Design of the
  2015 {{ChaLearn AutoML}} challenge,'' in \emph{Int. Joint Conf. Neural
  Networks ({IJCNN})}, Jul. 2015, pp. 1--8.

\bibitem{hutter_sequential_2011}
F.~Hutter, H.~H. Hoos, and K.~{Leyton-Brown},
  ``\BIBforeignlanguage{en}{Sequential {{Model}}-{{Based Optimization}} for
  {{General Algorithm Configuration}}},'' in
  \emph{\BIBforeignlanguage{en}{Learning and {{Intelligent Optimization}}}},
  ser. Lecture Notes in Computer Science.\hskip 1em plus 0.5em minus
  0.4em\relax {Springer, Berlin, Heidelberg}, Jan. 2011, pp. 507--523.

\bibitem{kotthoff_auto-weka_2017}
L.~Kotthoff, C.~Thornton, H.~H. Hoos, F.~Hutter, and K.~{Leyton-Brown},
  ``Auto-{{WEKA}} 2.0: {{Automatic}} model selection and hyperparameter
  optimization in {{WEKA}},'' \emph{J. Mach. Learn. Res.}, vol.~18, no.~25, pp.
  1--5, 2017.

\bibitem{hall_weka_2009}
M.~Hall, E.~Frank, G.~Holmes, B.~Pfahringer, P.~Reutemann, and I.~H. Witten,
  ``The {{WEKA Data Mining Software}}: {{An Update}},'' \emph{ACM SIGKDD
  Explorations Newsletter}, vol.~11, no.~1, pp. 10--18, Nov. 2009.

\bibitem{gandomi_handbook_2015}
A.~H. Gandomi, A.~H. Alavi, and C.~Ryan, Eds.,
  \emph{\BIBforeignlanguage{en}{Handbook of {{Genetic Programming
  Applications}}}}.\hskip 1em plus 0.5em minus 0.4em\relax {Springer
  International Publishing}, 2015.

\bibitem{graff_semantic_2017}
M.~Graff, E.~S. Tellez, H.~Jair~Escalante, and S.~{Miranda-Jim\'enez},
  ``\BIBforeignlanguage{en}{Semantic {{Genetic Programming}} for {{Sentiment
  Analysis}}},'' in \emph{\BIBforeignlanguage{en}{{{NEO}} 2015: {{Results}} of
  the {{Numerical}} and {{Evolutionary Optimization Workshop NEO}} 2015 Held at
  {{September}} 23-25 2015 in {{Tijuana}}, {{Mexico}}}}, ser. Studies in
  Computational Intelligence, O.~Sch\"utze, L.~Trujillo, P.~Legrand, and
  Y.~Maldonado, Eds.\hskip 1em plus 0.5em minus 0.4em\relax Cham: {Springer
  International Publishing}, 2017, pp. 43--65.

\bibitem{pawlak_semantic_2015}
T.~P. Pawlak, B.~Wieloch, and K.~Krawiec, ``Semantic {{Backpropagation}} for
  {{Designing Search Operators}} in {{Genetic Programming}},'' \emph{IEEE
  Transactions on Evolutionary Computation}, vol.~19, no.~3, pp. 326--340, Jun.
  2015.

\bibitem{ashlock_evolving_2015}
D.~Ashlock and J.~Tsang, ``Evolving fractal art with a directed acyclic graph
  genetic programming representation,'' in \emph{2015 {{IEEE Congress}} on
  {{Evolutionary Computation}} ({{CEC}})}, May 2015, pp. 2137--2144.

\bibitem{olson_automating_2016}
R.~S. Olson, R.~J. Urbanowicz, P.~C. Andrews, N.~A. Lavender, L.~C. Kidd, and
  J.~H. Moore, ``\BIBforeignlanguage{en}{Automating {{Biomedical Data Science
  Through Tree}}-{{Based Pipeline Optimization}}},'' in
  \emph{\BIBforeignlanguage{en}{Applications of {{Evolutionary Computation}}}},
  ser. Lecture Notes in Computer Science, vol. 9597.\hskip 1em plus 0.5em minus
  0.4em\relax {Springer, Cham}, Mar. 2016, pp. 123--137.

\bibitem{kordik2018discovering}
P.~Kord{\'\i}k, J.~{\v{C}}ern{\`y}, and T.~Fr{\`y}da, ``Discovering predictive
  ensembles for transfer learning and meta-learning,'' \emph{Machine Learning},
  vol. 107, no.~1, pp. 177--207, 2018.

\bibitem{de2017recipe}
A.~G. de~S{\'a}, W.~J.~G. Pinto, L.~O.~V. Oliveira, and G.~L. Pappa, ``Recipe:
  a grammar-based framework for automatically evolving classification
  pipelines,'' in \emph{European Conference on Genetic Programming}.\hskip 1em
  plus 0.5em minus 0.4em\relax Springer, 2017, pp. 246--261.

\bibitem{ghorbani2001stacked}
A.~A. Ghorbani and K.~Owrangh, ``Stacked generalization in neural networks:
  generalization on statistically neutral problems,'' in \emph{Neural Networks,
  2001. Proceedings. IJCNN'01. International Joint Conference on},
  vol.~3.\hskip 1em plus 0.5em minus 0.4em\relax IEEE, 2001, pp. 1715--1720.

\bibitem{kahn_topological_1962}
A.~B. Kahn, ``Topological {{Sorting}} of {{Large Networks}},''
  \emph{Communications of the ACM}, vol.~5, no.~11, pp. 558--562, Nov. 1962.

\bibitem{goldberg_comparative_1992}
D.~E. Goldberg and K.~Deb, ``A {{Comparative Analysis}} of {{Selection Schemes
  Used}} in {{Genetic Algorithms}},'' in \emph{Foundations of {{Genetic
  Algorithms}}}, G.~J. Rawlins, Ed.\hskip 1em plus 0.5em minus 0.4em\relax
  {Elsevier}, 1992, vol.~1, pp. 69--93.

\bibitem{vanschoren_openml_2014}
J.~Vanschoren, J.~N. {van Rijn}, B.~Bischl, and L.~Torgo, ``{{OpenML}}:
  {{Networked Science}} in {{Machine Learning}},'' \emph{ACM SIGKDD
  Explorations Newsletter}, vol.~15, no.~2, pp. 49--60, Jun. 2014.

\bibitem{dheeru_uci_2017}
D.~Dheeru and E.~Karra~Taniskidou, ``{{UCI Machine Learning Repository}},''
  {University of California, Irvine, School of Information and Computer
  Sciences}, Tech. Rep., 2017.

\bibitem{velez_balanced_2007}
D.~R. Velez, B.~C. White, A.~A. Motsinger, W.~S. Bush, M.~D. Ritchie, S.~M.
  Williams, and J.~H. Moore, ``\BIBforeignlanguage{en}{A {{Balanced Accuracy
  Function}} for {{Epistasis Modeling}} in {{Imbalanced Datasets Using
  Multifactor Dimensionality Reduction}}},''
  \emph{\BIBforeignlanguage{en}{Genetic Epidemiology}}, vol.~31, no.~4, pp.
  306--315, May 2007.

\end{thebibliography}

\end{document}